\documentclass[conference]{IEEEtran}
\IEEEoverridecommandlockouts
\usepackage{cite}
\usepackage{amsmath,amssymb,amsfonts}
\usepackage{algorithmic}
\usepackage{graphicx}
\usepackage{textcomp}
\usepackage{multirow}
\usepackage{xcolor}
\usepackage{makecell}

\usepackage{threeparttable}
\usepackage{url}
\usepackage{cite}
\usepackage{mathrsfs}
\usepackage{multirow}
\usepackage{booktabs}
\usepackage{stfloats}
\usepackage{enumitem}
\usepackage{bbding}
\usepackage{CJKutf8}

\usepackage{amssymb}
\usepackage[T1]{fontenc}

\def\BibTeX{{\rm B\kern-.05em{\sc i\kern-.025em b}\kern-.08em
    T\kern-.1667em\lower.7ex\hbox{E}\kern-.125emX}}
\begin{document}

\title{BLR-MoE: Boosted Language-Routing Mixture of Experts for Domain-Robust Multilingual E2E ASR\\
\thanks{-------------------------------------------------------------------------------------------}
\thanks{$*$ Corresponding author}
}


\author{\IEEEauthorblockN{Guodong Ma \quad\quad\quad\quad\quad}
\IEEEauthorblockA{\textit{Yidun  AI Lab \quad\quad\quad\quad\quad} \\
\textit{Netease} \quad\quad\quad\quad\quad \\
Hangzhou, China \quad\quad\quad\quad\quad\\
maguodong@corp.netease.com \quad\quad\quad\quad\quad}
\and
\IEEEauthorblockN{Wenxuan Wang \quad\quad\quad\quad\quad}
\IEEEauthorblockA{\textit{Yidun  AI Lab \quad\quad\quad\quad\quad} \\
\textit{Netease \quad\quad\quad\quad\quad}\\
Hangzhou, China \quad\quad\quad\quad\quad \\
wangwenxuan@corp.netease.com \quad\quad\quad\quad\quad}
\and
\IEEEauthorblockN{Lifeng Zhou}
\IEEEauthorblockA{\textit{Yidun  AI Lab} \\
\textit{Netease}\\
Hangzhou, China \\
hzzhoulifeng@corp.netease.com}
\and
\IEEEauthorblockN{Yuting Yang}
\IEEEauthorblockA{\textit{Yidun  AI Lab} \\
\textit{Netease}\\
Hangzhou, China \\
yangyuting04@corp.netease.com}
\and
\IEEEauthorblockN{\quad\quad\quad\quad Yuke Li}
\IEEEauthorblockA{\quad\quad\quad\quad\quad \textit{Yidun  AI Lab} \\
\textit{\quad\quad\quad\quad\quad  Netease}\\
\quad\quad\quad\quad\quad Hangzhou, China \\
\quad\quad\quad\quad\quad liyuke@corp.netease.com}
\and
\IEEEauthorblockN{\quad\quad\quad\quad\quad\quad~ Binbin Du$^*$}
\IEEEauthorblockA{\quad\quad\quad\quad\quad\quad\quad~ \textit{Yidun  AI Lab} \\
\quad\quad\quad\quad\quad\quad~ \textit{Netease}\\
\quad\quad\quad\quad\quad\quad\quad~ Hangzhou, China \\
\quad\quad\quad\quad\quad\quad\quad~ dubinbin@corp.netease.com}
}

\maketitle

\begin{abstract}
Recently, the Mixture of Expert (MoE) architecture, such as LR-MoE, is often used to alleviate the impact of language confusion on the multilingual ASR (MASR) task. However, it still faces language confusion issues, especially in mismatched domain scenarios. In this paper, we decouple language confusion in LR-MoE into confusion in self-attention and router. To alleviate the language confusion in self-attention, based on LR-MoE, we propose to apply attention-MoE architecture for MASR. In our new architecture, MoE is utilized not only on feed-forward network (FFN) but also on self-attention. In addition, to improve the robustness of the LID-based router on language confusion, we propose expert pruning and router augmentation methods. Combining the above, we get the boosted language-routing MoE (BLR-MoE) architecture. We verify the effectiveness of the proposed BLR-MoE in a 10,000-hour MASR dataset. 
\end{abstract}


\section{Introduction}
\label{sec:intro}

With the development of end-to-end (E2E) automatic speech recognition (ASR) systems, tasks in different scenarios \cite{graves2006connectionist,graves2012sequence,kim2017joint,speech_transformer,ma21_interspeech,yuhang_paper,enhance_language_prompt_frame,radford2023robust,wang23sa_interspeech,10447520,10096227,2024arXiv240606329L,10389662,10095133,10096812,song2024u2++} have achieved performance improvements. In multilingual ASR (MASR)
scenarios \cite{radford2023robust,LUPET_paper,2024arXiv240612611K,2024arXiv240606329L,wang23sa_interspeech,10096227,10447520,2024arXiv240602166Y, 10446800,2024arXiv240606619S, 2023arXiv230301037Z}
, researchers have tried to use many methods to improve the recognition performance, such as Mixture of Expert (MoE) \cite{wang23sa_interspeech,10096227,10447520}, multi-task learning \cite{radford2023robust,LUPET_paper}, and integrating language information \cite{2024arXiv240612611K,2024arXiv240606329L}, etc. However, the MoE-based architecture, such as LR-MoE \cite{wang23sa_interspeech}, still faces language confusion problems, especially in mismatched domain scenarios.

Firstly, the LR-MoE \cite{wang23sa_interspeech} only performs MoE on the Transformer's FFN (feed-forward network) module, and the attention part is still affected by language confusion. As we all know, the LR-MoE-based MASR \cite{wang23sa_interspeech} mainly uses the MoE module to form different sub-networks to reduce the recognition impact caused by language confusion. Recently, the pathways-based method \cite{10447373,10094300} a similar idea using sub-networks to improve MASR, but the sub-modules include FFN and attention. Moreover, in the language modeling task \cite{attmoe1,attmoe2}, the MoE-based attention has been proven effective.

Secondly, in the LR-MoE-based MASR, the accuracy of ASR highly depends on the accuracy of the router, which is affected by language confusion. The router is based on language identification (LID). Therefore, if the LID-based router makes an error, the data will be distributed to the wrong expert. 
However, mistakenly allocated data will be recognized as tokens of another language. Especially when there is a domain mismatch between the training and testing data, language confusion will often occur. It will affect the MASR performance.

Inspired by the above, based on LR-MoE \cite{wang23sa_interspeech}, we propose applying MoE to the attention module to alleviate confusion in self-attention. 
In addition, to solve the impact of language confusion in the LID-based router, we propose two solutions, namely router augmentation and expert pruning. Thus, the boosted LR-MoE architecture (BLR-MoE) is formed. As for router augmentation, it uses a stronger LID classifier and decouples the LID-based router and ASR model to make up for the shortcomings of BLR-MoE. Specifically, decoupling means that, given a trained BLR-MoE model, we will extract the LID-based router and use the audio-language pairs data (in real applications, such data is easier to obtain than the audio-text pair data) to tune only the LID-based router with very few parameters, so as not to affect the ASR part. By decoupling the router and ASR, we decouple the language confusion problem in MASR into the LID problem, that is, to resolve a relatively more difficult issue into a relatively simple one. Regarding expert pruning, given the prior language information, the BLR-MoE experts are pruned to reduce the impact of language confusion in the LID-based router and make a trained ASR model configurable for different application scenarios.

Our main contributions are as follows:
\begin{itemize}
    \item To alleviate confusion in self-attention and the LID-based router, based on LR-MoE, we propose attention-MoE, expert pruning, and router-augmentation for MASR, respectively, thus forming the BLR-MoE architecture.
    \item The BLR-MoE can enable the model to quickly adapt to different language and
     domain needs with the help of expert pruning and router augmentation, respectively.
    \item Our proposed BLR-MoE model has undergone rigorous evaluation on a 10,000-hour MASR dataset, demonstrating a significant 16.09\% relative reduction in WER over the LR-MoE model. Notably, the model exhibits substantial improvements of 3.98\% in in-domain and 19.09\% in out-of-domain scenarios. 
\end{itemize}


\section{Problem formulations and motivation}
\label{sec:format}

\begin{figure}[t]
\centering
 \centerline{\includegraphics[width=4.5cm]{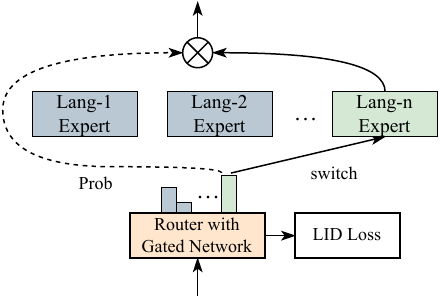}}
\caption{The processing of the MoE router }

\label{fig:mle}
\vspace{-1.5 em}
\end{figure}

In the MASR system, language information closely relates to the final ASR accuracy. Therefore, we can use the following formula to describe the modeling of the MASR system: 
\begin{eqnarray}
\setlength\tabcolsep{1.0pt}
\label{eq:multi_asr}
  \begin{aligned}
    \rm P ({\rm Y}| {\rm X)} = \rm \sum_{\rm L} \rm P ({\rm Y}, {\rm L} | {\rm X}) = \rm \sum_{\rm L} \rm P ({\rm Y} | {\rm L}, {\rm X}) \times \rm P ({\rm L} | {\rm X}),
  \end{aligned}
\end{eqnarray}
where $\rm X$ refers to the $\rm T$-length speech feature sequence, $\rm Y$ denotes label sequence, and the $\rm L$ represents the language label of $\rm X$.

In traditional Transformer-based MASR, all languages share the same Transformer module \cite{Transformer_raw,speech_transformer}. Due to differences in language and domain, this shared Transformer module has different degrees of fit for different languages and scenarios, resulting in different performances. In particular, due to language confusion, misrecognition of cross-language types will often occur, which will degrade the model performance.

The MASR architecture based on LR-MoE \cite{wang23sa_interspeech} is an effective solution to alleviate the problem of language confusion. Its core idea is to apply MoE to the FFN of Transformer to form different sub-networks for various languages, which make $\rm P ({\rm Y} | \rm {L}, \rm {X})$ in Eq.~\ref{eq:multi_asr} to be sparse, and cooperate with the LID-based router to distribute the data to be processed to the corresponding sub-networks for processing, as shown in Fig.~\ref{fig:mle}. The LID-based router can be regarded as an explicit modeling of $\rm P ( {\rm L} | {\rm X})$ in Eq.~\ref{eq:multi_asr}.

As for the LR-MoE-based MASR architecture, we need to consider two issues. First, it only applies MoE to FFN. However, recently, the pathways-based method \cite{10447373,10094300} shows that there seems to be a problem of language confusion in attention. In addition, attention-based MoE \cite{attmoe1,attmoe2} has verified in the language modeling task that it can further enhance the modeling ability of the Transformer. Second, the LID-based router is learned using the teacher-forcing training method. Therefore, each language-specific MoE module in our model is dedicated to learning ASR-related knowledge pertinent to its language. However, if an error occurs in the LID-based router, the data to be processed will be distributed to the wrong language expert, which will often cause errors in the final ASR recognition results. In particular, due to language confusion, incorrect distribution will often occur when there is a domain mismatch between the training and testing data.

\begin{figure}[t]
\centering
 \centerline{\includegraphics[width=8.0cm]{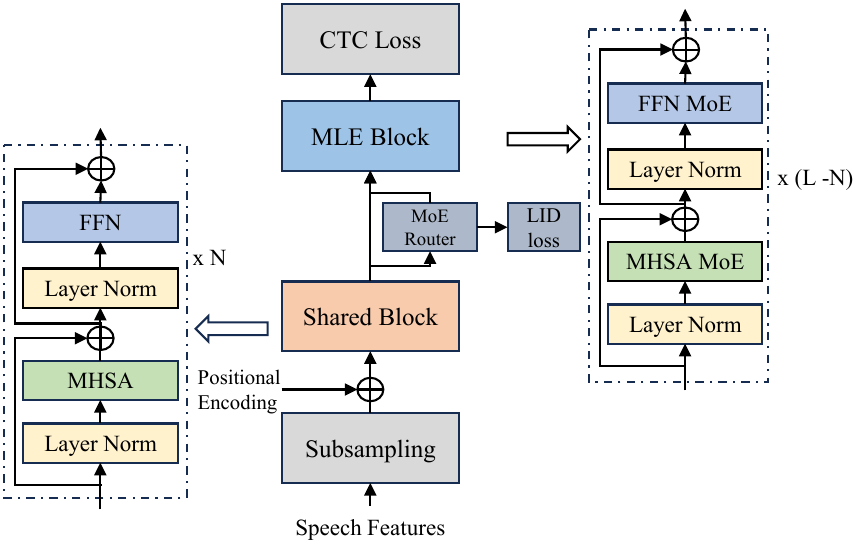}}
\vspace{-0.5 em}
\caption{The structure of the BLR-MoE Transformer Model. $N$ and $(L-N)$ are the number of layers of the shared block and the Mixture of Language Experts (MLE) block, respectively.}

\label{fig:arch}
\vspace{-1.5 em}
\end{figure}

\section{Proposed Method}
\label{sec:propose}

\subsection{Attention MoE}
\label{ssec:att_moe}
The LR-MoE-based MASR architecture only applies MoE on the Transformer's FFN module and can improve the model recognition performance. However, recent works show that sparse attention will also improve the modeling ability of the model \cite{attmoe1,attmoe2} and reduce language confusion in MASR \cite{10447373,10094300}. Therefore, based on the LR-MoE architecture \cite{wang23sa_interspeech}, we try to apply MoE to the self-attention module, as shown in Fig.~\ref{fig:arch}. We know that the self-attention module is mainly composed of four parts, namely key ($\rm \mathbf{K}$), query ($\rm \mathbf{Q}$), value ($\rm \mathbf{V}$), and output ($\rm \mathbf{O}$) module. We only take one head of multi-head self-attention (MHSA) as an example, which corresponds to four matrices $\rm \mathbf{W}_K^{h}$, $\rm \mathbf{W}_Q^{h}$, $\rm \mathbf{W}_V^{h}$, and $\rm \mathbf{W}_O^{h}$ respectively. When we choose to perform MoE on $\rm \mathbf{V}$ and $\rm \mathbf{O}$ in self-attention, given the intermediate representation $\rm H_{in}$, it can be expressed as follows:
\begin{eqnarray}
\label{eq:att_moe}
    \rm \mathbf{K} &=& \rm concat({H_{in} \mathbf{W}_K^{h}}) \\
    \rm \mathbf{Q} &=& \rm concat({H_{in} \mathbf{W}_Q^{h}}) \\
    \rm \mathbf{V} &=& \rm concat({H_{in} \mathbf{W}_V^{h,e}}) \\ 
    \rm \mathbf{A} &=& \rm softmax(\rm{\frac{\mathbf{QK^{T}}}{\sqrt{d_k}}}) \rm \mathbf{V} \\
    \rm \mathbf{O} &=& \rm {\mathbf{A} \mathbf{W}_O^{h,e}} \\
    \rm h &=& \{\rm{head1,...,headN}\} \\
    \rm e &\in& \rm E = \{\rm{e1,e2, ..., en}\}
\end{eqnarray}
where N denotes the number of attention heads, n refers to the number of supported language experts, and $\rm d_k$ is the dimension of the key. In addition, the language expert router for FFN and attention is shared.

By performing MoE on $\rm \mathbf{V}$ and $\rm \mathbf{O}$, the sub-network of the model has more independent modules, thereby reducing the impact of language confusion in MASR tasks and ultimately improving the overall performance of the model. Meanwhile, we conduct MoE experiments on different modules. In addition, it is different from the pathways-based method \cite{10447373,10094300}. In the pathways-based method, various languages have sub-networks of different sizes, which are related to the composition of the training data. However, in the MoE module of BLR-MoE, the sub-networks are the same size.

\begin{figure}[t]

\begin{minipage}[htb]{0.45\linewidth}
  \centering
  \centerline{\includegraphics[width=3.6cm]{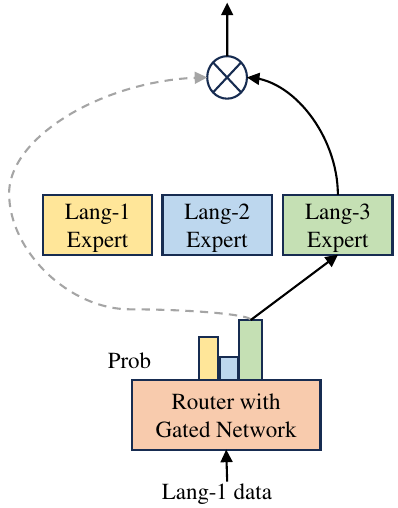}}
  \vspace{-0.3 em}
  \centerline{(a)} \medskip
\end{minipage}
%
\hfill
\begin{minipage}[htb]{0.45\linewidth}
  \centering
  \centerline{\includegraphics[width=4.cm]{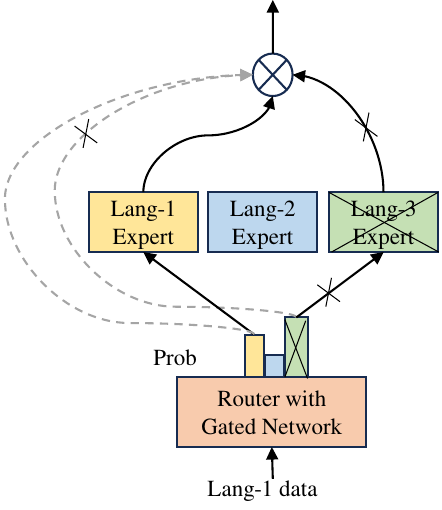}}
  \centerline{(b)} \medskip
\end{minipage}
%
\caption{Schematic diagram of MoE modules. (a) The raw MLE, (b) The MLE after expert pruning. \XSolid \quad indicates that the expert and path have been pruned.}
\vspace{-1.5 em}
\label{fig:moe_prun}
\end{figure}

\subsection{Router Augmentation}
\label{ssect:router_aug}

As shown in Eq.~\ref{eq:multi_asr} and Fig.~\ref{fig:arch}, the BLR-MoE-based architecture decouples MASR into two tasks: LID $\rm P(\rm L | \rm X)$ and ASR $\rm P(\rm{Y, L | X})$. At the training stage, ASR and LID tasks are jointly trained through multi-task learning. The previous MoE methods \cite{wang23sa_interspeech} adopted a simple linear layer to model the gated network. The modeling capability of its LID-gated network was largely based on the shared encoder with the ASR task. Therefore, our proposed method introduces a more complex LID module (such as TDNN \cite{tdnn}) into the gated network to obtain additional modeling capacity independent of the ASR branch, such as the role of the LID adapter. We can fine-tune the LID-gated network (freezing the remaining network weights) with the lower-cost audio-language pair data. In this way, the impact of language confusion on the MoE router can be alleviated so that LID adaptation can alleviate the domain mismatch in the MASR system. As a result, it improves the robustness of the MASR.

\subsection{Expert Pruning}
\label{ssec:expert_prune}
As discussed in Section~\ref{sec:format}, the MoE-based MASR will be affected by the LID-based router and its accuracy will be influenced by language confusion. In the inference stage, if a range of prior language information can be given,  we can use expert pruning to reduce language confusion in the LID-based router. To facilitate the description of expert pruning, we assume that the model supports only three languages. When language confusion happens, the data will be distributed to the error language expert, as shown in Fig.~\ref{fig:moe_prun} (a). If we know that the target does not contain lang-3, we can prune the expert corresponding to lang-3, as shown in Fig.~\ref{fig:moe_prun} (b). The purpose is to reduce errors in the LID-based router ($\rm P(\rm L| \rm X)$ in Eq.~\ref{eq:multi_asr}), thus reducing the impact of language confusion in the ASR part ($\rm P(\rm Y, \rm L | \rm X)$ in Eq.~\ref{eq:multi_asr}). By using expert pruning, without the need for additional model training, a trained BLR-MoE model can be configurable for different scenarios.




\section{Experiments}
\label{sec:expr}

\subsection{Datasets and experiment setup}

We collect more than 10,000 hours of multilingual data from different open datasets, as shown in Table~\ref{tab:used_data}. The Chinese data consists of Aishell-1 
\cite{bu2017aishell} and a part of WenetSpeech 
\cite{9746682}. The English is from Librispeech 
\cite{7178964} and the part of GigaSpeech 
\cite{chen21o_interspeech}. The Japanese is collected from the part of Reazonspeech 
\cite{reazonspeech}. The Arabic is the MGB2 dataset 
\cite{7846277}. In addition, we use the Commonvoice V18.0 dataset \cite{ardila-etal-2020-common} to simulate out-of-domain situations.
In particular, 
we only use the part of English Commonvoice data and use the speed perturbation \cite{speed-perturb} for other data.
In Table~\ref{tab:used_data}, M and K denote million and thousand, respectively. The vocabulary consists of 17,299 unique characters and BPE \cite{SentencePiece} tokens.

We use the CTC-only model with a $\mathbf{N_{Encoder}}$=12. The encoder is stacked Transformer blocks with an attention dimension of 512, 8 attention heads, and an FFN dimension of 2048. Following \cite{wang23sa_interspeech}, the MoE-based system consists of a 6-layer shared block and a 6-layer MLE block. The final loss combines ASR and LID using $\lambda_{lid}=0.3$. The TDNN-based router consists of a TDNN with 500 channels and a FFN. We use the Adam optimizer \cite{kingma2014adam} with a Transformer-lr scale of 1 and warmup steps of 25k to train 40 epochs on 6 Tesla A100 GPUs 40G. The dropout rate \cite{srivastava2014dropout} is 0.1 to prevent the model from over-fitting. In the training stage, we adopt a dynamic batch size strategy with a maximum batch size of 100K frames. In the decoding stage, we only use the CTC greedy search.

\vspace{-0.5 em}

\begin{table}[t] \footnotesize
\setlength\tabcolsep{4.5pt}
  \caption{The details of the used Datasets}
   \vspace{-1 em}
  \label{tab:used_data}
    \center
    \begin{tabular}{| c | c | c | c | c | c |} \hline
    
     & \textbf{EN} & \textbf{ZH} & \textbf{JA} & \textbf{AR} & \textbf{ALL} \\ \hline
    \multicolumn{6}{|c|}{\textbf{In-domain}} \\
    \hline
    \multicolumn{6}{|l|}{Train set} \\ 
    \hline
    {Dur. (Hrs)} & 3,000 & 3,281 & 3,077 & 1,200 & 10,558 \\ 
    \hline
    {Utt. (Num)} & 1.9 M & 4.7 M & 1.8 M & 0.4 M & 8.8 M \\ 
    \hline
    \multicolumn{6}{|l|}{Test set} \\ 
    \hline
    {Dur. (Hrs)} & 5.4 & 14.3 & 6 & 9.6 & 35.3 \\ 
    \hline
    {Utt. (Num)} & 2.6 K & 16.6 K & 3.5 K & 5.4 K & 28.1 K \\ 
    \hline
    \multicolumn{6}{|c|}{\textbf{The used Commonvoice datasets}} \\
    \hline
    \multicolumn{6}{|l|}{Train set with LID label only} \\ 
    \hline
    {Dur. (Hrs)} & 100 & 108.5 & 33.8 & 81.4 &  323.7 \\ 
    \hline
    {Utt. (Num)} & 82.5 K & 88.2 K & 30.3 K & 85.2 K & 286.2 K \\ 
    \hline
    \multicolumn{6}{|l|}{Test set} \\ 
    \hline
    {Dur. (Hrs)} & 27 & 17.5 & 8.9 & 12.6 & 63.6 \\ 
    \hline
    {Utt. (Num)} & 16.4 K & 10.6 K & 6.3 K & 10.4 K & 41.9 K \\
    \hline
    \end{tabular}
    \vspace{-1.5 em}
\end{table}

\subsection{Results}
\textbf{Main results: }
As shown in Table~\ref{tab:main_res}, compared with the Vallina CTC model (ID-0), the LR-MoE (ID-1) achieves significant performance improvement in all tests. However, we can observe that our proposed BLR-MoE (ID-2) shows a 16.09\% relative WER reduction over the LR-MoE baseline (ID-1). In particular, 19.09\% is improved in the Commonvoice tests. Moreover, the accuracy of the LID-based router shows a 9.41\% absolute improvement compared with ID-1. As shown in Table~\ref{tab:main_res}, when we reduce the LID tune, the model (ID-3) still exhibits a notable improvement of 7.36\% compared to ID-1. In particular, the model achieves 8.14\% improvements in Commonvoice tests. Finally, our model still relatively improves by 6.51\% when we remove Router Augmentation in BLR-MoE (ID-4), which shows that attention-MoE can alleviate the confusion in LR-MoE. In general, the results are consistent with our motivation that the confusion in LR-MoE still affects the performance and verifies the effectiveness of our proposed method. 
\\
\textbf{Attention MoE:}
From system ID-2 to ID-5 in Table~\ref{tab:res_moe_diff}, we can observe that, after performing MoE on different attention modules, the ASR performance is relatively improved by 3.24-7.31\% in Commonvoice tests compared with the LR-MoE baseline (ID-0). Moreover, the model parameter increases slightly. In addition, from ID-0 to ID-2, the results suggest that performing MoE on FFN and attention are complementary, which is consistent with our expectations.
\\
\textbf{Expert Pruning:}
Based on system ID-4 in Table~\ref{tab:main_res}, we use expert pruning to mitigate the impact of language confusion. From systems ID-2 to ID-5 in Table~\ref{tab:res_expert_pru}, we can see that if only one language information is given, expert pruning can improve the performance. However, we find that the improvement is different due to the difference in confusion relationship between different languages. Therefore, among the four languages, we can observe that Chinese and Japanese are more closely related, and Arabic and English are more likely to be confused. In addition, when we give more language information, the influence of language confusion will be smaller. If we only let the model recognize Arabic audio (ID-6), compared with ID-0, a relative improvement of 34.28\% is obtained in the Commonvoice Arabic test. These experiments show that language confusion seriously hinders the effect of the MoE-based MASR, and also demonstrate that expert pruning can effectively alleviate this impact and make the model configurable to different application scenarios.

\begin{table*}[t] \footnotesize
\renewcommand\tabcolsep{2.0pt}
 \centering
 \caption{ Results of proposed models and the baselines (WER/CER, \%). The numbers in brackets indicate the accuracy (\%) of the LID-based router. The k, q, v, and o represent the key, query, value, and output layer in self-attention, respectively. The f denotes the FFN.
 }
 \vspace{-0.5 em}
 \label{tab:main_res}
 \begin{tabular}{  c | l | c | c | c| c | c | c | c | c | c | c | c | c | c | c } \hline
 \multicolumn{1}{l|}{\multirow{2}{*}{\textbf{ID}}} &
    \multicolumn{1}{l|}{\multirow{2}{*}{\textbf{Model}}} &
    \multicolumn{1}{l|}{\multirow{2}{*}{\textbf{\makecell{LID\\ router}}}} &
    \multicolumn{1}{l|}{\multirow{2}{*}{\textbf{\makecell{MoE \\ Modules}}}} &
    \multicolumn{1}{c|}{\multirow{2}{*}{\textbf{\makecell{train/infer \\ Para. (M)}}}} &
    \multicolumn{5}{c|}{\textbf{In-domain test sets}}& 
    \multicolumn{5}{c|}{\textbf{Commonvoice test sets}} & 
    \multicolumn{1}{c}{\multirow{2}{*}{ \textbf{Avg$\_$all}}} \\ \cline{6-15}
    & & & & & \textbf{ \makecell{ZH}} & \textbf{EN} & \textbf{JA} & \textbf{AR} & \textbf{Avg} & \textbf{ZH} & \textbf{EN} & \textbf{JA} & \textbf{AR} &\textbf{Avg} & \\
    \hline
    \multicolumn{16}{l}{\small \textbf{Baselines}} \\
    \hline
    0 & Vallina CTC & / & / & 55.6/55.6 & 7.31 & 8.33 & 11.66 & 10.18 & 9.37 & 24.02 & 45.75 & 16.47 & 61.58 & 36.96 & 23.16 \\
    \hline
    1 & LR-MoE & FFN & f &  93.7/55.9 & \makecell{ 5.03 \\ (100)} & \makecell{ 6.52 \\ (100)} & \makecell{ 10.05 \\ (99.97)} & \makecell{ 8.57 \\ (99.98)} & \makecell{ 7.54 \\ \textbf{(99.99)}} & \makecell{ 19.73 \\ (99.37)} & \makecell{ 42.05 \\ (86.37)} & \makecell{ 14.09 \\ (97.66)} & \makecell{ 45.04 \\ (69.73)} & \makecell{ 30.23 \\ (88.28)} & \makecell{ 18.89 \\ (94.14)} \\
    \hline
    \multicolumn{16}{l}{\textbf{Our proposed}} \\
    \hline
    \footnotesize 2 & BLE-MoE & TDNN & o,v,f & 104.7/56.7 & \makecell{ 4.6 \\ (100)} & \makecell{ 6.17 \\ (99.96)} & \makecell{ 9.77 \\ (100)} & \makecell{ 8.42 \\ (99.96)} & \makecell{ \textbf{7.24} \\ (99.98)} & \makecell{ 18.89 \\ (99.67)} & \makecell{ 34.75 \\ (97.44)} & \makecell{ 12.78 \\ (99.23)} & \makecell{ 31.4 \\ (94.43)} & \makecell{ 24.46 \\ (97.69)} & \makecell{ \textbf{15.85} \\ (\textbf{98.84})} \\
    \hline
    3 & \quad- only LID tune & TDNN & o,v,f & 103.9/56.7 & \makecell{ 4.6 \\ (100)} & \makecell{ 6.17 \\ (99.92)} & \makecell{ 9.77 \\ (100)} & \makecell{ 8.42 \\ (99.96)} & \makecell{ \textbf{7.24} \\ (99.97)} & \makecell{ 18.92 \\ (99.45)} & \makecell{ 38.01 \\ (90.75)} & \makecell{ 12.8 \\ (99.02)} & \makecell{ 41.33 \\ (72.52)} & \makecell{ 27.77 \\ (90.44)} & \makecell{ 17.5 \\ (95.2)} \\
    \hline
    4 & \quad\quad- Router Augmentation & FFN & o,v,f & 103.1/55.9 & \makecell{ 4.73 \\ (100)} & \makecell{ 6.25 \\ (99.96)} & \makecell{ 9.74 \\ (99.97)} & \makecell{ 8.46 \\ (99.96)} & \makecell{ 7.30 \\ (99.97)} & \makecell{ 19.01 \\ (99.32)} & \makecell{ 38.11 \\ (90.17)} & \makecell{ 12.85 \\ (98.76)} & \makecell{ 42.12 \\ (71.01)} & \makecell{ 28.02 \\ (89.82)} & \makecell{ 17.66 \\ (94.89)} \\    
    \hline
 \end{tabular}
 \vspace{-1.5 em}
\end{table*}

\begin{table}[t] \footnotesize
\renewcommand\tabcolsep{1pt}
 \centering
 \caption{Results on different attention MoE modules (WER/CER, \%)}
 \vspace{-0.2 em}
 \label{tab:res_moe_diff}
 \begin{tabular}{  c | l | c | c| c | c | c | c | c } \hline
    \textbf{ID} &
    \textbf{Model} &
    \textbf{\makecell{MoE \\ Modules}} &
    \textbf{\makecell{ Para. (M)}} &
    \textbf{ZH} & \textbf{EN} & \textbf{JA} & \textbf{AR} & \textbf{Avg} \\
    \hline
    \multicolumn{9}{l}{\small \textbf{Baseline}} \\
    \hline
    0 & LR-MoE & f &  93.7 & \makecell{ 19.73 \\ (99.37)} & \makecell{ 42.05 \\ (86.37)} & \makecell{ 14.09 \\ (97.66)} & \makecell{ 45.04 \\ (69.73)} & \makecell{ 30.23 \\ (88.28)} \\
    \hline
    \multicolumn{9}{l}{\textbf{Attention MoE}} \\
    \hline
    1 & BLR-MoE & o & 60.6 & \makecell{ 22.05 \\ (99.01)} & \makecell{ 44.25 \\ (86.1)} & \makecell{ 13.98 \\ (98.73)} & \makecell{ 49.29 \\ (62.76)} & \makecell{ 32.39 \\ (86.65)} \\
    \hline
    2 & BLR-MoE & o,f & 98.4 & \makecell{ 19.14 \\ (99.13)} & \makecell{ 39.89 \\ (90.42)} & \makecell{ 13.15 \\ (98.59)} & \makecell{ 44.21 \\ (69.24)} & \makecell{ 29.1 \\ (89.35)} \\
    \hline
    3 & BLR-MoE & o,v,f & 103.1 & \makecell{ \textbf{19.01} \\ (99.32)} & \makecell{ \textbf{38.11} \\ (90.17)} & \makecell{ \textbf{12.85} \\ (98.76)} & \makecell{ \textbf{42.12} \\ \textbf{(71.01)}} & \makecell{ \textbf{28.02} \\ \textbf{(89.82)}} \\
    \hline
    4 & BLR-MoE & v,k,q,f & 107.8 & \makecell{ 19.2 \\ (99.17)} & \makecell{ 39.64 \\ \textbf{(90.64)}} & \makecell{ 13.28 \\ \textbf{(98.84)}} & \makecell{ 44.89 \\ (69.55)} & \makecell{ 29.25 \\ (89.63)} \\
    \hline
    5 & BLR-MoE & o,v,k,q,f & 112.5 & \makecell{ 19.19 \\ \textbf{(99.45)}} & \makecell{ 39.64 \\ (89.86)} & \makecell{ 13.19 \\ (98.73)} & \makecell{ 43.13 \\ (69.66)} & \makecell{ 28.79 \\ (89.43)} \\
    \hline
 \end{tabular}
 \vspace{-1.2 em}
\end{table}
\begin{table}[t] \footnotesize
\renewcommand\tabcolsep{2.6pt}
 \centering
 \caption{Results on Expert Pruning (WER/CER, \%)}
 \vspace{-0.3 em}
 \label{tab:res_expert_pru}
 \begin{tabular}{  c | l | c | c| c | c | c | c } \hline
    \textbf{ID} &
    \textbf{Model} &
    \textbf{\makecell{MoE \\ Modules}} &
    \textbf{\makecell{ Experts \\ Pruning}} &
    \textbf{ZH} & \textbf{EN} & \textbf{JA} & \textbf{AR} \\
    \hline
    \multicolumn{8}{l}{\small \textbf{Baseline}} \\
    \hline
    0 & LR-MoE & f &  / & \makecell{ 19.73 \\ (99.37)} & \makecell{ 42.05 \\ (86.37)} & \makecell{ 14.09 \\ (97.66)} & \makecell{ 45.04 \\ (69.73)} \\
    \hline
    \multicolumn{8}{l}{\textbf{Expert Pruning}} \\
    \hline
    1 & BLR-MoE & o,v,f & / & \makecell{ 19.01 \\ (99.32)} & \makecell{ 38.11 \\ (90.17)} & \makecell{ 12.85 \\ (98.76)} & \makecell{ 42.12 \\ (71.01)} \\
    \hline
    2 & BLR-MoE & o,v,f & ZH & \makecell{ - } & \makecell{ 37.12 \\ (92.55)} & \makecell{ 12.72 \\ (99.69)} & \makecell{ 41.93 \\ (72.12)} \\
    \hline
    3 & BLR-MoE & o,v,f & EN & \makecell{ 19.01 \\ (99.32)} & \makecell{ - } & \makecell{ 12.8 \\ (99.02)} & \makecell{ 37.39 \\ (77.12)} \\
    \hline
    4 & BLR-MoE & o,v,f & JA & \makecell{ \textbf{18.91} \\ \textbf{(99.84)}} & \makecell{ \textbf{36.91} \\ \textbf{(93.27)}} & \makecell{ - } & \makecell{ 41.01 \\ (72.84)} \\
    \hline
    5 & BLR-MoE & o,v,f & AR & \makecell{ 19.01 \\ (99.32)} & \makecell{ 38.01 \\ (90.85)} & \makecell{ 12.58 \\ (98.76)} & \makecell{ - } \\
    \hline
    6 & BLR-MoE & o,v,f & ZH,EN & \makecell{ - } & \makecell{ - } & \makecell{ \textbf{12.6} \\ \textbf{(99.96)}} & \makecell{ 33.73 \\ (85.71)} \\
    \hline
    7 & BLR-MoE & o,v,f & ZH,EN,JA & \makecell{ - } & \makecell{ - } & \makecell{ - } & \makecell{ \textbf{27.68} \\ \textbf{(100)}} \\
    \hline
 \end{tabular}
 \vspace{-1.87 em}
\end{table}

\vspace{-0.468 em}
\section{Conclusions}
\label{sec:conclusion}

In this paper, we decouple the modeling process of multilingual ASR (MASR) into two tasks: LID and ASR. After decoupling, we find that the LR-MoE-based MASR still has language confusion in self-attention and the LID-based router. To alleviate these confusions, based on LR-MoE, we propose attention-MoE, expert pruning, and router-augmentation, respectively, thus forming the BLR-MoE architecture. After conducting experiments on the 10,000-hour MASR dataset, compared to the LR-MoE baseline, we find that BLR-MoE has achieved a significant improvement in the out-of-domain test and a slight improvement in the in-domain test.


\clearpage
\bibliographystyle{IEEEbib}
\bibliography{refs}

\begin{thebibliography}{10}

\bibitem{graves2006connectionist}
Alex Graves, Santiago Fern{\'a}ndez, Faustino Gomez, and J{\"u}rgen Schmidhuber,
\newblock ``Connectionist temporal classification: labelling unsegmented sequence data with recurrent neural networks,''
\newblock in {\em ICML}, 2006, pp. 369--376.

\bibitem{graves2012sequence}
Alex Graves,
\newblock ``Sequence transduction with recurrent neural networks,''
\newblock {\em arXiv preprint arXiv:1211.3711}, 2012.

\bibitem{kim2017joint}
Suyoun Kim, Takaaki Hori, and Shinji Watanabe,
\newblock ``Joint ctc-attention based end-to-end speech recognition using multi-task learning,''
\newblock in {\em ICASSP 2017}. IEEE, 2017, pp. 4835--4839.

\bibitem{speech_transformer}
Linhao Dong, Shuang Xu, and Bo~Xu,
\newblock ``Speech-transformer: A no-recurrence sequence-to-sequence model for speech recognition,''
\newblock in {\em ICASSP 2018}, 2018, pp. 5884--5888.

\bibitem{ma21_interspeech}
Guodong Ma, Pengfei Hu, Jian Kang, Shen Huang, and Hao Huang,
\newblock ``{Leveraging Phone Mask Training for Phonetic-Reduction-Robust E2E Uyghur Speech Recognition},''
\newblock in {\em Proc. Interspeech 2021}, 2021, pp. 306--310.

\bibitem{yuhang_paper}
Yuhang Yang, Haihua Xu, Hao Huang, Eng~Siong Chng, and Sheng Li,
\newblock ``Speech-text based multi-modal training with bidirectional attention for improved speech recognition,''
\newblock in {\em ICASSP 2023}, 2023, pp. 1--5.

\bibitem{enhance_language_prompt_frame}
Song Li, Yongbin You, Xuezhi Wang, Ke~Ding, and Guanglu Wan,
\newblock ``Enhancing multilingual speech recognition through language prompt tuning and frame-level language adapter,''
\newblock in {\em ICASSP 2024}, 2024, pp. 10941--10945.

\bibitem{radford2023robust}
Alec Radford, Jong~Wook Kim, Tao Xu, Greg Brockman, Christine McLeavey, and Ilya Sutskever,
\newblock ``Robust speech recognition via large-scale weak supervision,''
\newblock in {\em ICML}. PMLR, 2023, pp. 28492--28518.

\bibitem{wang23sa_interspeech}
Wenxuan Wang, Guodong Ma, Yuke Li, and Binbin Du,
\newblock ``{Language-Routing Mixture of Experts for Multilingual and Code-Switching Speech Recognition},''
\newblock in {\em Proc. INTERSPEECH 2023}, 2023, pp. 1389--1393.

\bibitem{10447520}
Thomas~Palmeira Ferraz, Marcely Zanon~Boito, Caroline Brun, and Vassilina Nikoulina,
\newblock ``Multilingual distilwhisper: Efficient distillation of multi-task speech models via language-specific experts,''
\newblock in {\em ICASSP 2024}, 2024, pp. 10716--10720.

\bibitem{10096227}
Yoohwan Kwon and Soo-Whan Chung,
\newblock ``Mole : Mixture of language experts for multi-lingual automatic speech recognition,''
\newblock in {\em ICASSP 2023}, 2023, pp. 1--5.

\bibitem{2024arXiv240606329L}
Wei Liu, Jingyong Hou, Dong Yang, Muyong Cao, and Tan Lee,
\newblock ``A parameter-efficient language extension framework for multilingual asr,''
\newblock in {\em Interspeech 2024}, 2024, pp. 3929--3933.

\bibitem{10389662}
Guodong Ma, Wenxuan Wang, Yuke Li, Yuting Yang, Binbin Du, and Haoran Fu,
\newblock ``Lae-st-moe: Boosted language-aware encoder using speech translation auxiliary task for e2e code-switching asr,''
\newblock in {\em 2023 IEEE Automatic Speech Recognition and Understanding Workshop (ASRU)}, 2023, pp. 1--8.

\bibitem{10095133}
Qianying Liu, Zhuo Gong, Zhengdong Yang, Yuhang Yang, Sheng Li, Chenchen Ding, Nobuaki Minematsu, Hao Huang, Fei Cheng, Chenhui Chu, and Sadao Kurohashi,
\newblock ``Hierarchical softmax for end-to-end low-resource multilingual speech recognition,''
\newblock in {\em ICASSP 2023}, 2023, pp. 1--5.

\bibitem{10096812}
Saierdaer Yusuyin, Hao Huang, Junhua Liu, and Cong Liu,
\newblock ``Investigation into phone-based subword units for multilingual end-to-end speech recognition,''
\newblock in {\em ICASSP 2023}, 2023, pp. 1--5.

\bibitem{song2024u2++}
Xingchen Song, Di~Wu, Binbin Zhang, Dinghao Zhou, Zhendong Peng, Bo~Dang, Fuping Pan, and Chao Yang,
\newblock ``U2++ moe: Scaling 4.7 x parameters with minimal impact on rtf,''
\newblock {\em arXiv preprint arXiv:2404.16407}, 2024.

\bibitem{LUPET_paper}
Wei Liu, Jingyong Hou, Dong Yang, Muyong Cao, and Tan Lee,
\newblock ``Lupet: Incorporating hierarchical information path into multilingual asr,''
\newblock in {\em Interspeech 2024}, 2024, pp. 3979--3983.

\bibitem{2024arXiv240612611K}
Yosuke Kashiwagi, Hayato Futami, Emiru Tsunoo, Siddhant Arora, and Shinji Watanabe,
\newblock ``Rapid language adaptation for multilingual e2e speech recognition using encoder prompting,''
\newblock in {\em Interspeech 2024}, 2024, pp. 2900--2904.

\bibitem{2024arXiv240602166Y}
Saierdaer {Yusuyin}, Te~{Ma}, Hao {Huang}, Wenbo {Zhao}, and Zhijian {Ou},
\newblock ``{Whistle: Data-Efficient Multilingual and Crosslingual Speech Recognition via Weakly Phonetic Supervision},''
\newblock {\em arXiv e-prints}, p. arXiv:2406.02166, June 2024.

\bibitem{10446800}
Yerbolat Khassanov, Zhipeng Chen, Tianfeng Chen, Tze~Yuang Chong, Wei Li, Lu~Lu, and Zejun Ma,
\newblock ``Extending multilingual asr to new languages using supplementary encoder and decoder components,''
\newblock in {\em ICASSP 2024 - 2024 IEEE International Conference on Acoustics, Speech and Signal Processing (ICASSP)}, 2024, pp. 10586--10590.

\bibitem{2024arXiv240606619S}
Zheshu Song, Jianheng Zhuo, Yifan Yang, Ziyang Ma, Shixiong Zhang, and Xie Chen,
\newblock ``Lora-whisper: Parameter-efficient and extensible multilingual asr,''
\newblock in {\em Interspeech 2024}, 2024, pp. 3934--3938.

\bibitem{2023arXiv230301037Z}
Yu~{Zhang}, Wei {Han}, James {Qin}, Yongqiang {Wang}, Ankur {Bapna}, Zhehuai {Chen}, Nanxin {Chen}, Bo~{Li}, Vera {Axelrod}, Gary {Wang}, Zhong {Meng}, Ke~{Hu}, Andrew {Rosenberg}, Rohit {Prabhavalkar}, Daniel~S. {Park}, Parisa {Haghani}, Jason {Riesa}, Ginger {Perng}, Hagen {Soltau}, Trevor {Strohman}, Bhuvana {Ramabhadran}, Tara {Sainath}, Pedro {Moreno}, Chung-Cheng {Chiu}, Johan {Schalkwyk}, Fran{\c{c}}oise {Beaufays}, and Yonghui {Wu},
\newblock ``{Google USM: Scaling Automatic Speech Recognition Beyond 100 Languages},''
\newblock {\em arXiv e-prints}, p. arXiv:2303.01037, Mar. 2023.

\bibitem{10447373}
Jiamin Xie, Ke~Li, Jinxi Guo, Andros Tjandra, Yuan Shangguan, Leda Sari, Chunyang Wu, Junteng Jia, Jay Mahadeokar, and Ozlem Kalinli,
\newblock ``Dynamic asr pathways: An adaptive masking approach towards efficient pruning of a multilingual asr model,''
\newblock in {\em ICASSP 2024}, 2024, pp. 12201--12205.

\bibitem{10094300}
Mu~Yang, Andros Tjandra, Chunxi Liu, David Zhang, Duc Le, and Ozlem Kalinli,
\newblock ``Learning asr pathways: A sparse multilingual asr model,''
\newblock in {\em ICASSP 2023}, 2023, pp. 1--5.

\bibitem{attmoe1}
R{\'o}bert {Csord{\'a}s}, Piotr {Pi{\k{e}}kos}, Kazuki {Irie}, and J{\"u}rgen {Schmidhuber},
\newblock ``{SwitchHead: Accelerating Transformers with Mixture-of-Experts Attention},''
\newblock {\em arXiv e-prints}, p. arXiv:2312.07987, Dec. 2023.

\bibitem{attmoe2}
Xun {Wu}, Shaohan {Huang}, Wenhui {Wang}, and Furu {Wei},
\newblock ``{Multi-Head Mixture-of-Experts},''
\newblock {\em arXiv e-prints}, p. arXiv:2404.15045, Apr. 2024.

\bibitem{Transformer_raw}
Ashish Vaswani, Noam Shazeer, Niki Parmar, Jakob Uszkoreit, Llion Jones, Aidan~N Gomez, \L~ukasz Kaiser, and Illia Polosukhin,
\newblock ``Attention is all you need,''
\newblock in {\em Advances in Neural Information Processing Systems}, I.~Guyon, U.~V. Luxburg, S.~Bengio, H.~Wallach, R.~Fergus, S.~Vishwanathan, and R.~Garnett, Eds. 2017, vol.~30, Curran Associates, Inc.

\bibitem{tdnn}
A.~Waibel, T.~Hanazawa, G.~Hinton, K.~Shikano, and K.J. Lang,
\newblock ``Phoneme recognition using time-delay neural networks,''
\newblock {\em IEEE Transactions on Acoustics, Speech, and Signal Processing}, vol. 37, no. 3, pp. 328--339, 1989.

\bibitem{bu2017aishell}
Hui Bu, Jiayu Du, Xingyu Na, Bengu Wu, and Hao Zheng,
\newblock ``Aishell-1: An open-source mandarin speech corpus and a speech recognition baseline,''
\newblock in {\em 2017 O-COCOSDA}. IEEE, 2017, pp. 1--5.

\bibitem{9746682}
Binbin Zhang, Hang Lv, Pengcheng Guo, Qijie Shao, Chao Yang, Lei Xie, Xin Xu, Hui Bu, Xiaoyu Chen, Chenchen Zeng, Di~Wu, and Zhendong Peng,
\newblock ``Wenetspeech: A 10000+ hours multi-domain mandarin corpus for speech recognition,''
\newblock in {\em ICASSP 2022}, 2022, pp. 6182--6186.

\bibitem{7178964}
Vassil Panayotov, Guoguo Chen, Daniel Povey, and Sanjeev Khudanpur,
\newblock ``Librispeech: An asr corpus based on public domain audio books,''
\newblock in {\em ICASSP 2015}, 2015, pp. 5206--5210.

\bibitem{chen21o_interspeech}
Guoguo Chen, Shuzhou Chai, Guan-Bo Wang, Jiayu Du, Wei-Qiang Zhang, Chao Weng, Dan Su, Daniel Povey, Jan Trmal, Junbo Zhang, Mingjie Jin, Sanjeev Khudanpur, Shinji Watanabe, Shuaijiang Zhao, Wei Zou, Xiangang Li, Xuchen Yao, Yongqing Wang, Zhao You, and Zhiyong Yan,
\newblock ``{GigaSpeech: An Evolving, Multi-Domain ASR Corpus with 10,000 Hours of Transcribed Audio},''
\newblock in {\em Proc. Interspeech 2021}, 2021, pp. 3670--3674.

\bibitem{reazonspeech}
Yue Yin, Daijiro Mori, and Seiji Fujimoto,
\newblock ``Reazonspeech: A free and massive corpus for japanese asr,''
\newblock in {\em NLP 2023}, 2023.

\bibitem{7846277}
Ahmed Ali, Peter Bell, James Glass, Yacine Messaoui, Hamdy Mubarak, Steve Renals, and Yifan Zhang,
\newblock ``The mgb-2 challenge: Arabic multi-dialect broadcast media recognition,''
\newblock in {\em SLT 2016}, 2016, pp. 279--284.

\bibitem{ardila-etal-2020-common}
Rosana Ardila, Megan Branson, Kelly Davis, Michael Kohler, Josh Meyer, Michael Henretty, Reuben Morais, Lindsay Saunders, Francis Tyers, and Gregor Weber,
\newblock ``Common voice: A massively-multilingual speech corpus,''
\newblock in {\em Proceedings of the Twelfth Language Resources and Evaluation Conference}, Marseille, France, May 2020, pp. 4218--4222, European Language Resources Association.

\bibitem{speed-perturb}
T.~Ko, Vijayaditya Peddinti, D.~Povey, and S.~Khudanpur,
\newblock ``Audio augmentation for speech recognition,''
\newblock in {\em INTERSPEECH}, 2015.

\bibitem{SentencePiece}
Taku {Kudo} and John {Richardson},
\newblock ``{SentencePiece: A simple and language independent subword tokenizer and detokenizer for Neural Text Processing},''
\newblock {\em arXiv e-prints}, p. arXiv:1808.06226, Aug. 2018.

\bibitem{kingma2014adam}
Diederik~P Kingma and Jimmy Ba,
\newblock ``Adam: A method for stochastic optimization,''
\newblock {\em arXiv preprint arXiv:1412.6980}, 2014.

\bibitem{srivastava2014dropout}
Nitish Srivastava, Geoffrey Hinton, Alex Krizhevsky, Ilya Sutskever, and Ruslan Salakhutdinov,
\newblock ``Dropout: a simple way to prevent neural networks from overfitting,''
\newblock {\em The journal of machine learning research}, vol. 15, no. 1, pp. 1929--1958, 2014.

\end{thebibliography}

\end{document}